\documentclass[runningheads]{llncs}

\usepackage{eccv}

\usepackage{eccvabbrv}

\usepackage{graphicx}
\usepackage{booktabs}

\usepackage[accsupp]{axessibility}  

\usepackage{amsmath}
\usepackage{amssymb}
\usepackage{algorithm}
\usepackage{algorithmic}
\usepackage{multirow} 
\usepackage{makecell}
\usepackage{float}

\usepackage{hyperref}

\usepackage{orcidlink}
\usepackage{subcaption}

\begin{document}
\begin{sloppypar}

\title{G-EvoNAS: Evolutionary Neural Architecture Search Based on Network Growth} 

\titlerunning{Abbreviated paper title}

\author{Juan Zou \and
Weiwei Jiang \and
Yizhang Xia\and
Yuan Liu\and
Zhanglu Hou}

\authorrunning{F.~Author et al.}

\institute{Xiangtan University, Yuhu District, Xiangtan, Hunan, China
\email{202221632973@smail.xtu.edu.cn}
\\
}

\maketitle

\begin{abstract}
  The evolutionary paradigm has been successfully applied to neural network search(NAS) in recent years. Due to the vast search complexity of the global space, current research mainly seeks to repeatedly stack partial architectures to build the entire model or to seek the entire model based on manually designed benchmark modules. The above two methods are attempts to reduce the search difficulty by narrowing the search space. To efficiently search network architecture in the global space, this paper proposes another solution, namely a computationally efficient neural architecture evolutionary search framework based on network growth (G-EvoNAS). The complete network is obtained by gradually deepening different Blocks. The process begins from a shallow network, grows and evolves, and gradually deepens into a complete network, reducing the search complexity in the global space. Then, to improve the ranking accuracy of the network, we reduce the weight coupling of each network in the SuperNet by pruning the SuperNet according to elite groups at different growth stages. The G-EvoNAS is tested on three commonly used image classification datasets, CIFAR10, CIFAR100, and ImageNet, and compared with various state-of-the-art algorithms, including hand-designed networks and NAS networks. Experimental results demonstrate that G-EvoNAS can find a neural network architecture comparable to state-of-the-art designs in 0.2 GPU days.
\end{abstract}

\section{Introduction}
\label{sec:intro}

In recent years, neural networks have experienced great success in solving various challenging tasks \cite{kenton2019bert,he2016deep,huang2017densely}, such as image recognition \cite{han2019attribute,he2016deep}, object detection \cite{girshick2015fast} and semantic segmentation \cite{he2017mask}. Since the performance of deep neural networks majorly depends on their architecture, an increasing number of research efforts in the deep learning community are devoted to designing novel architectures, such as DenseNet \cite{huang2017densely}, ResNet \cite{he2016deep}, and GoogleNet \cite{szegedy2015going}. However, the design of novel network architectures strongly relies on the knowledge and experience of human experts and may necessitate multiple attempts before meaningful results are achieved \cite{he2016deep}. Neural Architecture Search (NAS) is regarded as an effective method to automate the design of task-specific neural network architectures \cite{tan2019mnasnet}, which enables interested users to quickly develop their own CNN app without sufficient engineering experience and domain expertise. It has been shown that NAS can implement competitive DNN architectures and human experts and discover new state-of-the-art architectures \cite{zhang2020one}.

Present-day NAS methods mainly search local architectures (Cell) and then stack the searched Cells to build the entire model architecture or use artificially designed modules as basic search units to search the global neural network model. The former method simply repeatedly stacks Cells to build a model, resulting in the structure of each model layer being the same. Indeed, the functions of each neural network layer are different, so the resulting architecture could be more optimal. The global optimal solution cannot be obtained by reducing the complexity of the final model structure in exchange for fast convergence of the search process. The alternative method of directly searching the entire neural network architecture \cite{guo2020single} generally uses artificially designed modules such as MobileNetV2 \cite{sandler2018mobilenetv2} as the basic building block. Using artificially designed block structures as basic units, this method performs well. Search methods targeting block structures are more straightforward than searching for neural network architectures in the global space, but the search space is greatly restricted.

Despite the fact that local search methods and methods based on manually designed modules can achieve excellent performance, the potential performance of globally operated search methods isn't overlooked~\cite{xu2022dnas}. The objective of this study is to search global neural network models more efficiently. A growth-centric network method is proposed. Through this method, a diverse inter-layer Block neural network model architecture can be obtained without directly searching the global space in a less time-consuming manner, thus enabling a quick and effective search of the entire model architecture.

\begin{figure}[!t]
  \centering
    \includegraphics[scale=0.3]{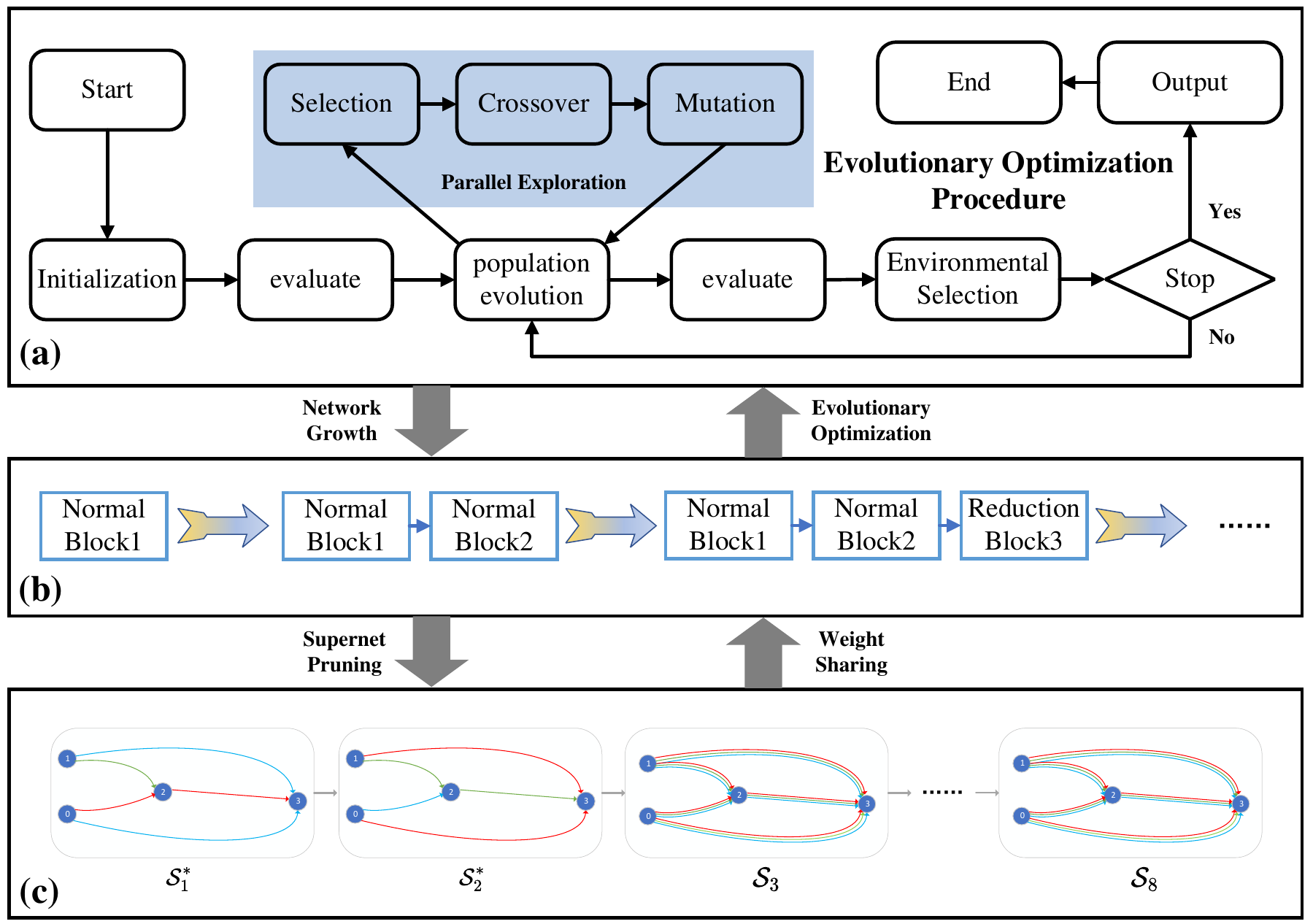}
  \caption{Overall framework of G-EvoNAS.(a) Evolutionary search process. During the search process of population evolution, only the latest growing Block will be searched. (b) Network growth process, the network length will increase by one Block at each growth stage. (c) Supernet pruning, the SuperNet corresponding to the current growth stage will be searched. The block in is pruned. Figure (c) shows an example of pruning two blocks.}
  \label{fig:1}
\end{figure}

In this paper, we propose an evolutionary neural architecture search framework based on network growth(G-EvoNAS). Specifically, we divide the evolution process into multiple growth stages and search for block structures in the network one by one in a growing manner. At the end of each growth stage, the candidate network in the current population is retained as the initial network for the next evolutionary stage, and the process is repeated. process until a sufficient number of blocks are found to form a complete network. Since only a single block is searched at each growth stage during the evolution process, the huge complexity caused by searching directly in the global space can be avoided. Lastly, concerning accelerated assessment, the weight-sharing strategy of the SuperNet is utilized to evaluate the performance of all progeny individuals without training. This not only reduces computational costs, but also, to alleviate the weight coupling of each network within the SuperNet, we preserve the superior individuals at each evolutionary stage from the population to conduct phasic pruning of the SuperNet, reducing the quantity of candidate networks within the SuperNet and enhancing the ranking accuracy of the network.An overview of our approach is shown in Figure 1.

The main contributions of our work are summarized as follows:
\begin{itemize}
  \item An alternative approach to reduce the global search complexity of NAS , i.e.,progressively searching each block structure in the candidate network and allowing the network to grow from a single block to a comprehensive configuration. 

  \item The weight sharing of the SuperNet is used to evaluate the candidate networks, while pruning the SuperNet based on the elite population during the network growth process. The weight coupling between individual subnets is alleviated by reducing the number of candidate networks in the SuperNet. The accuracy of network ranking is improved while accelerating individual performance evaluation.

  \item Our algorithm searches for competitive network architectures in a relatively short time. Using only 0.2 GPU days on the CIFAR10 and CIFAR100 datasets, the final network architectures achieve 97.52\% and 83.38\% classification accuracy. Afterwards, we transfer the network structure searched on CIFAR10 directly to the ImageNet dataset and find that G-EvoNAS can achieve 75.5\% classification accuracy. 

\end{itemize}

\section{Related Work}
\label{sec:RelatedWork}
\subsection{EA-based NAS}
The development of neural networks via evolutionary algorithms has garnered considerable interest. With the initial application of evolutionary algorithms in neural network architecture search (NEAT), remarkable success has been achieved \cite{levine2004non}. Evolutionsssary algorithms have been broadly utilized in the optimization of neural network parameters and hyperparameters \cite{fernandes2020automatic,guo2020single,real2017large}, and currently, an abundance of research is concentrating on ENAS\cite{girshick2015fast,he2017mask}. 
\cite{real2017large} unveiled what was likely the inaugural large-scale application of a simple evolutionary algorithm, which returns networks matching the human-designed models. The evolutionary process is described in detail in \cite{real2019regularized} and the concept of age is introduced.The framework of our paper is based on evolutionary algorithms, where common genetic operators, tournament selection strategies, and environmental selection strategies match or even outperform their RL baselines in terms of speed and accuracy \cite{real2019regularized}. However, since a large number of candidate models often need to be trained from scratch for evaluation, most of these methods are still computationally intensive.

\subsection{SMBO-based NAS}

Several previous works have explored the Sequence Model-Based Optimization (SMBO). \cite{negrinho2017deeparchitect} involves Monte Carlo Tree Search (MCTS). The sequential model-based optimization approach \cite{hutter2011sequential} amplifies the efficiency of MCTS by unraveling a predictive model that directs node expansion. Other relevant works include the likes of\cite{mendoza2016towards,stanley2002evolving,zoph2016neural,grosse2012exploiting}, which investigate MLPs instead of CNNs, the application of incremental methods in topology, the augmentation of the layer count, and the search for the syntactically specific area of the latent factor model respectively. Works such as \cite{cortes2017adanet,huang2018learning} progressively blossom CNNs sequentially; \cite{liu2018progressive} gradually probe the entire network by adding nodes to the Cell and hastening performance evaluation via the surrogate model; and \cite{li2020improving}, which employs progressive search to alleviate the posterior decay. 


\subsection{Pruning SuperNet}

The paradigm of Neural Architecture Search (NAS) \cite{he2016deep,tan2019mnasnet} is typically marked by high computational costs. To address this challenge, route improvements in search performance by sharing weights across varying models have been attempted \cite{liu2018darts,pham2018efficient}. 
These methods, which leverage shared weights across a hyper-parameterized network (hypergraph) that contains every model, can be further categorized into two groups. 

The first, continuous relaxation methods \cite{liu2018darts}, co-optimizes the weights and architectural selection factors of the SuperNet through gradient descent. However, the optimization of these selection factors introduces biases between sub-models inevitably. As sub-models with initial low performance will garner less training, they tend to be outperformed by other models. These methods heavily rely on the initial state, making it challenging to achieve optimal architecture. The second, one-shot methods \cite{guo2020single,brock2017smash,chu2021fairnas}, which guarantee the equitable treatment of all sub-models. After the SuperNet is trained via path loss or path sampling, the sub-models are sampled and assessed using weights inherited from the SuperNet. 

Other relatable works include ASAP \cite{noy2020asap} and XNAS\cite{nayman2019xnas}, models that introduce pruning during the training process of over-parameterized networks to improve NAS efficiency. Similar to these methods, we commence from an over-parameterized network, then reduce the search area to harvest an optimized architecture, and further improve the model's ranking.

\section{Methodology}
\label{sec:meth}


In this section, we introduce G-EvoNAS in detail from the aspects of search space, search strategy and acceleration method. First, based on the existing micro search space, we propose a block search space, which increases the diversity of network modules. An evolutionary search strategy based on network growth is proposed in the block search space, which further reduces the search complexity. Finally, through SuperNet pruning, the accuracy of network ranking in the SuperNet is improved.

\subsection{Block-wise Search Space}


The primary focus of this section is to present the specifics of the search space employed in our study. As illustrated in Figure 1, the overall network architecture comprises of varying Blocks. All prospective networks possess a fixed outer structure. Each Block receives direct input from the preceding Block (as demonstrated in the figure) and input from its prior Block. Each component block exhibits distinct characteristics, yet, they generally can be categorized into two types: normal Blocks and downsampling Blocks. These vary in that their internal structure is different and that the internal stride of the Reduction Block lessens the magnitude of the image size, whereas the Normal Block retains it. From the illustration, it is evident that the Blocks located at the 1/3 and 2/3 markers of the completed network are designated as Reduction Blocks, while the remainder are Normal Blocks. The ultimate goal of architecture striving is to unearth the structure of diverse blocks.

We design new efficient networks by utilizing different topologies, allowing each Block to have its own structure. Although the block construction process is the same as in \cite{liu2018darts}, the construction of the entire network is different. Specifically, the chunked search space is an extended version of the micro-search space proposed in \cite{zoph2016neural} and\cite{pham2018efficient}. To handle intermediate information, our method designs a new convolutional block after downsampling instead of repeating a normal block multiple times. This scheme overcomes the limitation of repeating the same normal block throughout the network, since the same kind of block may not be optimal for feature maps of different resolutions.




\begin{figure*}[!t]
  \centering
    \includegraphics[scale=0.55]{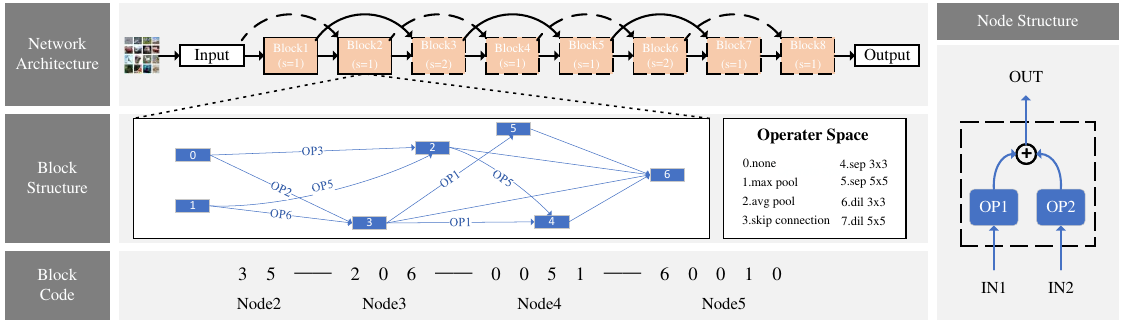}
  \caption{Top: A design network architecture consisting of 8 blocks. Middle and bottom: Block 2 and its block code example. A block consists of 4 hidden nodes, 2 source nodes (node 0 and node 1) and one output node (node 6). Right side: nodes represent two inputs that are given an output by an add operation after the corresponding operation.}
  \label{fig:2}
\end{figure*}

\begin{algorithm}
	\renewcommand{\algorithmicrequire}{\textbf{Input:}}
	\renewcommand{\algorithmicensure}{\textbf{Output:}}
	\caption{The overall framework of G-EvoNAS}
	\label{alg1}
	\begin{algorithmic}[1]
    \REQUIRE $B$(max num blocks), $G$(number of evolutionary iterations per block), $S$(SuperNet), $P_{num}$(population number), $D_{train}$ (train set), $D_{val} $(validataion set), $E_w$(warm up epoch), $E_s$(epoch interval of population evolution)
		\STATE Warm-up$(S, D_{train}, E_w)$
    \STATE Network $ \leftarrow \left \{\emptyset \right \}  $   //Set of candidate structuresk
    \STATE $ {P_0} \leftarrow$ InitializePopulation$(P_{num}, Network) $
    \FOR{$b \leftarrow 1: B$}
      \STATE $t \leftarrow 0$
      \IF {$b > 1 $}
          \STATE PruneSupernet$(S, P_t^b)$
      \ENDIF
      
      \STATE GrowthNetworks$(P_t^b, Block_{1:P_{num}}^b)$
      \WHILE{$t < G$}
        \STATE $Q_t^b \leftarrow $Generate offspring from $P_t^b$ using genetic operators
        \STATE $P_t^b \leftarrow P_t^b \cup Q_t^b $
        \STATE TrainSupernet$(S, D_{train}, E_s, P_t^b)$
        \STATE Evaluate$(P_t^b, S, D_{val})$ //Evaluate the fitness of individuals in $P_t^b$ on $D_{valid}$
        \STATE $P_{t+1}^b \leftarrow $Using pNSGA\uppercase\expandafter{\romannumeral3} Select from $P_t^b$
        \STATE $t \leftarrow t+1$

      \ENDWHILE
      
    \ENDFOR
    \ENSURE $P_t^B$ as the final result.

	\end{algorithmic}  
\end{algorithm}

\subsection{Evolution Search}
Many previous methods only search in the space of local cells, or search in the space of CNN based on handcrafted models. Although this is a simpler and more direct approach, it gives up the possibility of finding a globally optimal model. While we believe that it is difficult to directly navigate an exponentially large search space, especially when starting the process without valid model knowledge, an alternative approach to reducing search complexity should be taken rather than simply narrowing down search space.

To address this issue, we employ a concept akin to\cite{liu2018progressive} to optimize $Block_{1:N}$ through a sequential model. This approach significantly minimizes the complexity of our problem. Particularly, we parcel out the network search procedure into numerous stages for scrutinization and incrementally search the block structure within each aspirant network in the population one by one. Over the course of this procedure, the structure of the candidate network evolves from an individual block. To finalize the network, we initially fabricate the potential network structure of a singular block (i.e., composed of 1 block) in the premier growth stage and incorporate them into the population. Simultaneously, we train and assess all models within the population in a SuperNet (concurrently), developing for G generations until we achieve a converged population. Thereafter, in the second stage, we expand the population by appending a possible second block structure for each Model. At each evolutionary stage, we separately search for a corresponding block in all candidate networks in the population and extend it to the matching candidate network. Subsequently, we randomly sample the remaining blocks to form a complete candidate network to evaluate their potential. Those deemed elite Individuals proceed to the next generation population. At the climax of each evolutionary stage, the candidate network in the current population is retained as the base network for the subsequent evolutionary stage, and the process is iterated. For specific procedures, refer to Algorithm 1.

\noindent {\bf Genome Encoding.}
This paper provides a hybrid real-coded representation of the network structure under constraints, unlike most previous NAS work which uses binary adjacency matrix coding to represent the network structure. Each element of the adjacency matrix uses a decimal integer to encode the operation. Afterwards, by row-major vectorization (that is, flattening) the adjacency matrix, we can encode any architecture in the search space.See Figure 2 (bottom) for specific representation.


\noindent {\bf Crossover and Mutation Operator.}
In the population-based search method, potential population individuals can be used in parallel to build a new network structure through crossover. This paper designs a constrained uniform crossover operator that can cross the connections and operations of the network at the same time. In order to enhance the diversity of the population (with different network architectures) and the ability to jump out of the local optimum, we design two main mutations. The difference between the two mutations is that connection mutations are used to change the connection method, while combinatorial operation mutations are used to change the operations inside the node.Figure 3 provides examples of crossover and mutation operators.

\noindent {\bf Potential Assessment of Candidates.}
In the process of evolution, we will gradually eliminate disadvantaged individuals in the population through environmental selection pNSGA\uppercase\expandafter{\romannumeral3} algorithm\cite{yang2020cars}. The factor we consider here is the potential of each individual architecture at this time. We define the potential of an individual as $ P(\mathcal N)$, containing the expected validation accuracy of the model and the expected model size:
\begin{equation}\label{eq1}
 P(\mathcal N)=E_{\mathcal N \in \{\mathcal N|\mathcal N_i=Block^b,\forall b \leq B\}}(Acc(\mathcal N),Size(\mathcal N)) .
\end{equation}
\noindent Included in this process, the anticipated validation accuracy for model $\mathcal N$ is denoted by $Acc(\mathcal N)$. The prospective model size for model $\mathcal N$ is represented by $Size(\mathcal N)$. With uniform sampling of valid models, each chosen model uses a mini-batch to calculate its validation accuracy. The average accuracy across each sampled model over the full validation set is used to estimate the expected validation accuracy for model $\mathcal N$. This process also incorporates the average model size of each sampled model to predict the expected model size for model $\mathcal N$. We discerned that given a sufficient count of sampled models, the potential of candidate models remains quite consistent and distinguishable among the models. 

\subsection{Supernet training and pruning}

\noindent {\bf Inaccurate evaluation in NAS.} The most time-consuming operation in the NAS process is performance evaluation. This is because to accurately evaluate the candidate network, it is necessary to train it from scratch and test its real performance until it converges. However, the performance of this method The evaluation cost is high and obviously unacceptable. In order to speed up the process of performance evaluation, recent works \cite{liu2018darts,pham2018efficient} do not recommend training candidate networks from scratch until convergence, because they believe that it is unwise to directly discard the previously trained network weights , which can be reused to train different candidate architectures simultaneously by using shared network parameters. Specifically, they define the search space $\mathcal S$ as a SuperNet, such that each candidate network $\mathcal N$ is its subnet. Let $\mathcal W$ represent the weight parameter of the SuperNet, and the training of the SuperNet is as follows:
\begin{equation}\label{eq2}
 \mathcal W^*= \mathop{\arg\min}_{\mathcal W} L_{train}(\mathcal S,\mathcal W;X,Y) ,   
\end{equation}
where $L_{train}$ represents the training loss, while $X$ and $Y$ represent the input image and the corresponding label. Then, the performance of different candidate architectures is evaluated by sharing the optimal weight parameter $\mathcal W^*$ obtained after training. However, the optimal weight parameter $\mathcal W^*$ is optimal for the SuperNet, but it does not mean that $\mathcal W^*$ is the optimal weight parameter for each candidate architecture, because the subnet has not been fairly and fully trained, and the evaluation using $\mathcal W^*$ cannot Rank candidate models correctly because the search space is often large. Inaccuracies in assessment contribute to the ineffectiveness of existing NAS.

\noindent {\bf Supernet pruning.} \cite{chu2021fairnas,li2020improving}  show that when the search space is small and all candidates are fully and fairly trained, the evaluation can be accurate. In order to improve the accuracy of evaluation, we prune the SuperNet into blocks. Specifically, let N denote SuperNet. We divide $\mathcal S$ into $B$ blocks according to the depth of the SuperNet and have:
\begin{equation}\label{eq3}
\mathcal S = \mathcal S_1 \circ \dots \mathcal S_i\circ \mathcal S_{i+1} \dots \circ \mathcal S_B ,
\end{equation}
where $S_{i} \circ \mathcal S_{i+1}$ represents the $(i + 1)$-th block initially connected to the $i$-th block in the SuperNet. We then prune each block of the SuperNet in turn using:
\begin{equation}\label{eq4}
\mathcal S_{1:i}^* = \bigcup\limits_{j=1}^{P_{num}}\mathcal N_j  \quad (\mathcal N_j \in P_G^i)\quad i = 2,3 \dots, B, 
\end{equation}
where $\mathcal S_{1:i}^*$ represents the SuperNet after pruning from the $1$st block to the $i$-th block, $\mathcal N_j$ represents the subnet in $\mathcal A_{1:i}^*$, and $P_G^i$ is the final population at the $i$-th growth stage. For more accurate evaluation subnet, merge the two parts of the SuperNet and then train the pruned SuperNet:
\begin{equation}\label{eq5}
\mathcal W^*= \mathop{\arg\min}_{\mathcal W} L_{train}(\mathcal S_{1:i}^*,\mathcal S_{(i+1):B};X,Y),    
\end{equation}
where $\mathcal S_{i:B}$ represents the unpruned remaining blocks in the SuperNet. In order to ensure the effective reduction of weight-sharing search space in block NAS, the pruning rate is analyzed as follows. Let $C$ denotes the number of candidate networks for the $i$-th block. Then the size of the search space of the $i$-th block is $C_i$, $\forall i \in [1, B]$; the size of the search space $\mathcal S$ is $\displaystyle\prod_{i = 1}^{B}C_i$. This shows that the size of the weight-sharing search space decreases exponentially:
\begin{equation}\label{eq6}
Dropout\ rate = \displaystyle\prod_{j = 1}^{i}P_{num} \cdot \displaystyle\prod_{j = i+1}^{B}C_j /( \displaystyle\prod_{i = 1}^{B}C_i ).
\end{equation}
In our experiments, the weight-sharing search space in the SuperNet is significantly reduced , ensuring that each candidate architecture $\mathcal N  \in \mathcal S$ is fully optimized.
\section{Experiments Results}
\label{sec:results}
In this section, we execute experiments on CIFAR10 and CIFAR100 datasets. Having acquired the optimal neural network configuration with the best performance on CIFAR10 and CIFAR100, we implemented it for the ImageNet classification task. The experimental outcomes substantiate the generalization capability of the network we have constructed. 

\begin{table*}[t]
  \centering
  \caption{PERFORMANCE COMPARISON OF G-EvoNAS WITH COMPETITORS IN TERMS OF THE CLASSIFICATION ACCURACY (\%), NUMBER OF PARAMETERS AND THE SEARCH COST (GPU DAYS) ON CIFAR10 AND CIFAR100 DATASETS}
  \label{table1}
  \begin{tabular}{c|ccc|ccc|c}
    \hline\hline
    \multirow{2}{*}{Architecture} & 
    \multicolumn{3}{c|}{CIFAR10} &
    \multicolumn{3}{c|}{CIFAR100} &
    \multirow{2}{*}{Search Level}\\
    \cline{2-7}
    &Acc (\%) &P (M)&GDs&Acc (\%) &P (M)&GDs& \\
    \hline
    MobileNetV2\cite{sandler2018mobilenetv2}   &94.56          &2.1&-      &77.09         &2.1&-&manual\\
    \hline                   
    NASNet-A\cite{zoph2016neural}      &97.35          &3.2&1800   &82.19         &3.2&1800&Repeated Cell\\
    PNAS\cite{liu2018progressive}          &96.59          &3.2&225    &82.37         &3.2&225&Repeated Cell\\
    ENAS\cite{pham2018efficient}          &97.11          &4.6&0.5    &81.09         &4.6&0.5&Repeated Cell\\
    AmoebaNet-A[44]\cite{real2019regularized}   &96.66          &3.1&3150   &81.07         &3.1&3150&Repeated Cell\\
    SNAS\cite{xie2018snas}          &97.15          &2.8&1.5    &-                     &-&-&Repeated Cell\\
    DARTS\cite{liu2018darts}         &97.14          &3.3&1      &82.46         &3.3&1&Repeated Cell\\
    PDARTS\cite{chen2019progressive}        &97.50          &3.4&0.3    &84.08         &3.6&0.3&Repeated Cell\\
    PC-DARTS\cite{xu2019pc}      &97.43          &3.6&0.1    &82.89         &3.6&0.1&Repeated Cell\\
    NSGANetV1-A3\cite{lu2020multiobjective}  &97.78          &2.2&27     &82.77         &2.2&27&Repeated Cell\\
    CARS-I\cite{yang2020cars}        &97.38          &3.6&0.4    &82.72         &3.6&0.4&Repeated Cell\\
    MOEA-PS\cite{xue2023neural}       &97.23          &3.0&2.6    &81.03         &5.8&5.2&Repeated Cell\\
    \hline         
    Proxyless NAS\cite{cai2018proxylessnas}  &\textbf{97.92}&5.7&1500   &-                     &-&-&Artificial Block\\
    NSGA-Net\cite{lu2019nsga}      &97.25          &3.3&4      &79.26         &3.3&8&Artificial Block\\
    CNN-GA\cite{sun2020automatically}        &96.78          &2.9&35     &79.47         &3.3&1800&Artificial Block\\
    AE-CNN\cite{sun2019completely}        &95.30          &2.0&27     &79.15         &5.4&36&Artificial Block\\
    EPCNAS-C\cite{huang2022particle}      &96.93          &1.2&1.1    &81.67         &1.3&1.1&Artificial Block\\
    \hline          
    SI-EvoNAS\cite{zhang2020efficient}     &97.31          &1.8&0.5   &\textbf{84.30}&3.3&0.8&Complete Network\\
    DNAS\cite{xu2022dnas}          &97.32          &3.3&0.4   &82.53         &4.6&0.6&Complete Network\\
    \hline         
    \textbf{G-EvoNAS}&97.52          &3.2&0.2    &83.38         &3.3&0.2&Growth Network\\
    \hline\hline

  \end{tabular}
\end{table*}

\subsection{Results on CIFAR10 and CIFAR100}
\noindent {\bf Training Details.} Throughout the search process, we partitioned the original 50,000 training images into sets of 40,000 for training purposes and 10,000 for validation. We constructed a compact SuperNet composed of 16 channels and 8 regional blocks. The count of nodes within each Block is fixed at 7. Further, we incorporated the label smoothing method \cite{szegedy2016rethinking} to avoid overfitting. To optimize the SuperNet, we utilized SGD. The momentum was set at 0.9, and the weight decay was recorded as $3\times 10^{-4}$. The primary learning phase adheres to a cosine annealing schedule \cite{loshchilov2016sgdr} wherein the learning rate was reduced from 0.025 to 0 progressively. We specified the batch size of the training set and validation as 256 and 512 respectively.

For comprehensive CIFAR10 training, we elongate the stacking architecture by adjusting the number of blocks and initial number of channels to 20 and 36 respectively. The derived architecture will be trained from its inception for 600 epochs employing all training images, and test performance will be measured on the test set, utilizing the SGD optimizer with a momentum of 0.9. The initial learning rate is designated as 0.025 and a weight decay of $3\times 10^{-4}$ is applied. We employ a cosine annealing scheme \cite{loshchilov2016sgdr} to progressively reduce the learning rate until it reaches 0. The batch size is standardized at 96, and the specifications of other parameters remain consistent with \cite{liu2018darts}.

\noindent {\bf Comparison with State-of-the-art. } Table 1 delineates the comparative analysis with state-of-the-art classification results on CIFAR10 and CIFAR100. In contrast to the manually engineered network, MobileNet V2 \cite{sandler2018mobilenetv2}, the models we have sought after have considerably outperformed their networks. Adhering to a fair approach, we also juxtaposed with alternative neural architecture search methodologies. This included stacked Cell-based search algorithms such as NASNet-A\cite{zoph2016neural}, PNAS\cite{liu2018progressive}, AmoebaNet-A\cite{real2019regularized}, DARTS\cite{liu2018darts}, PDARTS\cite{chen2019progressive}, PC-DARTS\cite{xu2019pc}, CARS-I\cite{yang2020cars}, and others. Upon contrast with other networks, G-EvoNAS delivers competitive results on CIFAR10 and CIFAR100 with equivalent or lower search costs, although NSGANetV1-A3 \cite{lu2020multiobjective} displays superior classification precision with fewer parameters than G-EvoNAS. However, the search expense of G-EvoNAS is significantly lower (only comprising 0.2 GD, substantially lower than 27 GD). When considered against manual block-based search algorithms, G-EvoNAS performs superiorly to networks with comparable parameters like NSGA-Net \cite{lu2019nsga}, CNN-GA \cite{sun2020automatically}, AE-CNN \cite{sun2019completely}, and demonstrates superior performance on CIFAR10 and CIFAR100 with a search cost merely 0.2 GD. Although Proxyless NAS \cite{cai2018proxylessnas} exhibits better classification precision than GENet on CIFAR10, GENet deploys fewer parameters (3.2M \textless 5.7M). Upon contrast with the algorithm that conducts a direct search of the complete network, G-EvoNAS utilizes less time for seeking superior networks on CIFAR10. While it falls short of SI-EvoNAS\cite{zhang2020efficient} on CIFAR100 by 0.92\%, the search time of G-EvoNAS is merely 1/4 of it. The system thus discovers networks with similar model sizes and performances.


\begin{table*}[t]
  \centering
  \caption{COMPARISON BETWEEN G-EvoNAS AND OTHER NAS METHODS IN TERMS OF CLASSIFICATION ACCURACY (\%), THE NUMBER OF PARAMETERS, AND THE CONSUMED SEARCH COST ON THE IMAGENET DATASET}
  \label{table2}
  \begin{tabular}{c|cc|c|c|c}
    \hline\hline
    \multirow{2}{*}{Architecture} & 
    \multicolumn{2}{c|}{Test Acc(\%)} &
    \multirow{2}{*}{Params(M)} &
    \multirow{2}{*}{\makecell[c]{Search Cost\\ (GPU-days)}} &
    \multirow{2}{*}{Search Level}\\
    \cline{2-3}
    & Top-1 &Top-5& &  & \\
    \hline
    MobileNetV2\cite{sandler2018mobilenetv2}  &72.0   &90.4   &3.4    &-      &manual\\
    ShuffleNetV2\cite{ma2018shufflenet} &74.9   &90.1   &7.4    &-      &manual\\
    \hline
    AmoebaNet-A\cite{real2019regularized}      & 74.5           & 92.0  & 5.1   & 3150  & Repeated Cell \\
    NASNet-A\cite{zoph2016neural}         & 74             & 91.6  & 5.3   & 1800  & Repeated Cell \\
    PNAS\cite{liu2018progressive}             & 74.2           & 91.9  & 5.1   & 224   & Repeated Cell \\
    DARTS\cite{liu2018darts}            & 73.3           & 91.3  & 4.7   & 4     & Repeated Cell \\
    PDARTS\cite{chen2019progressive}           & 75.6           & 92.6  & 4.9   & 0.3   & Repeated Cell \\
    SNAS\cite{xie2018snas}             & 72.7           & 90.8  & 4.3   & 1.5   & Repeated Cell \\
    ENAS\cite{pham2018efficient}             & 74.3           & 91.9  & 5.1   & 0.5   & Repeated Cell \\
    CARS-I\cite{yang2020cars}           & 75.2           & 92.5  & 5.1   & 0.4   & Repeated Cell \\
    NSGANetV1-A3\cite{lu2020multiobjective}     & \textbf{76.2}  & 93.0    & 5.0     & 27    & Repeated Cell \\
    MOEA-PS\cite{xue2023neural}          & 73.6           & 91.5  & 4.7   & 2.6   & Repeated Cell \\
    \hline         
    Proxyless NAS\cite{cai2018proxylessnas}    & 75.1           & 92.5  & 7.1   & 8.3   & Artificial Block \\
    MNasNet-A2\cite{tan2019mnasnet}       & 75.6           & 92.7  & 4.8   & -     & Artificial Block \\
    EPCNAS-C2\cite{huang2022particle}        & 72.9           & 91.5  & 3.0  & 1.2  & Artificial Block \\
    FBNet-C\cite{wu2019fbnet}          & 74.9           & -     & 5.5   & 9     & Artificial Block \\
    \hline         
    SI-EvoNAS\cite{zhang2020efficient}        & 75.8           & 92.5 & 4.7   & 0.5   & Complete Network \\
    \hline         
    \textbf{G-EvoNAS}   & 75.5           & 92.5 & 4.6   & 0.2   & Growth Network \\
    \hline\hline         
         
  \end{tabular}
\end{table*}

\subsection{Results on ImageNet}
\noindent {\bf Training Details. }We adhere to the common practice of training networks on ImageNet \cite{simonyan2014very}. We initially preprocess every image and secure a random crop of the image of dimensions $224 \times 224$. The trimmed image segments are subsequently randomly flipped in a horizontal manner, after which the channel mean is deducted and divided by the channel standard deviation. We employ marginally less aggressive scale augmentation for lesser models, akin to the modifications used in \cite{howard2017mobilenets}. In the course of testing, we alter the size of the input image to $256\times 256$ and deploy a center crop of size $224\times 224$ as network input.

We utilize the SGD optimizer with a momentum of 0.9 and a weight decay of $3\times 10^{-5}$. The initial learning rate is set to 0.1, and a cosine annealing scheme \cite{loshchilov2016sgdr} is deployed to decrement the learning rate until it zeroes out. We use NVIDIA A100-SXM4-80GB for training with a batch size of 256. A maximum of 250 epochs are trained. Label smoothing of 0.1 and dropout with probability 0.3 were also amalgamated during training.

\noindent {\bf Comparison with State-of-the-art. } We assess the transferability of the search architecture by tutoring it on the ILSVRC2012 dataset. The outcomes on ILSVRC2012 are depicted in Table 2. G-EvoNAS exceeds manually crafted models, inclusive of MobileNetV2 \cite{sandler2018mobilenetv2} and ShuffleNetV2 \cite{ma2018shufflenet}, along with preceding state-of-the-art NAS models. This is particularly the case for NASNet\cite{zoph2016neural} and AmoebaNet\cite{real2019regularized}, which are emblematic of RL-based and EA-based methodologies, respectively. They consume more GPUs and days than our method. Upon contrast with alternative advanced NAS models, the stacked Cell-based search algorithms NSGANetV1-A3 \cite{lu2020multiobjective} and PDARTS \cite{chen2019progressive} outrank G-EvoNAS in performance by 0.7\% and 0.1\% respectively. However, the model sizes of both are larger than that of G-EvoNAS, and the search duration of G-EvoNAS is shorter (0.2\textless 0.3\textless 27). Upon contrast with PNAS\cite{liu2018progressive} , GENet utilizes lesser parameters (4.6M \textless 5.1M) and augmentation in top-1 and top-5 classification. The rates underwent an augmentation of 1.3\% and 0.6\% respectively. In the artificial block-based search algorithm and the algorithm that directly searches the complete network, the comparison of the MNasNet-A2 \cite{tan2019mnasnet} and SI-EvoNAS \cite{zhang2020efficient} networks, albeit the performance is marginally superior to G-EvoNAS, finds our model size to be smaller. The search cost is also less, especially for MNasNet-A2 \cite{tan2019mnasnet}, our search cost is exponentially reduced. Therefore, the G-EvoNAS search process is efficient and exhibits massive potential in searching for exceptional models.

\subsection{Ablation Study}
In this section, we investigate the effect of hyperparameters and discuss the effectiveness of our Supernet pruning and search methods.

\noindent {\bf Impact of hyperparameters} We examine the influence of hyperparameters on our methodology on the CIFAR10 dataset. The hyperparameters encompass the interval training generation $E_s$, the populace size $P_{num}$, and the stage evolution generation $G$. We endeavored setting $E_s$ to 50 and 100, maintaining $P_{num}$ = 10 and $G$ = 30. Experiments unveiled that the two resulting models did not exhibit sizable differences in reaching top-1 accuracy (less than 0.05\%). Consequently, we select the interval training generation $E_s$ to be 50 in our experiments to conserve computing resources. Pertaining to the effects of $P_{num}$ and $G$, we exemplify the results in Figure 3 (a). It is observable that the top-1 accuracy of the model discovered by G-EvoNAS doesn't inflate with the increase of $P_{num}$ and $G$. Consequently we select $P_{num}$ = 10, $G$ = 30 in our experiments for the potential of superior performance. We do not discern significant advancements when elevating these two hyperparameters any further in our experiment.
\begin{figure}[htbp]    
  \centering
\subfloat[]   
  {
      \label{fig:subfig1}\includegraphics[width=0.4\textwidth]{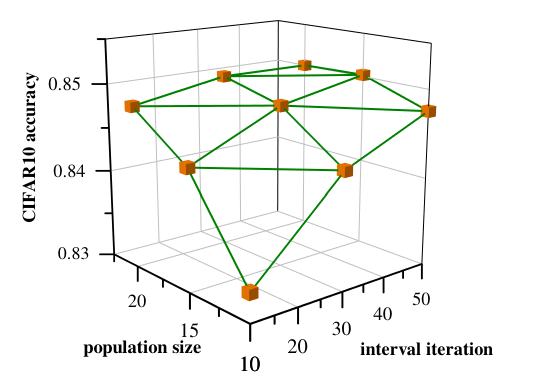}
  }
  \subfloat[]
  {
      \label{fig:subfig2}\includegraphics[width=0.55\textwidth]{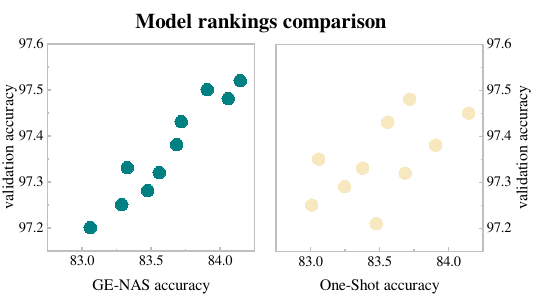}
  }
  \caption{(a)Hyperparametric analysis on CIFAR10. We chose population sizes of 10,15, and 20 and number of growth phase iterations of 10, 30, and 50. (b)Comparison of model rankings for G-EvoNAS (left) and One-Shot (right).}    
  \label{fig:subfig_1} 
\end{figure}

\noindent {\bf Effectiveness of our search method} Given the complexity in conveniently seeking the optimal solution in the enormous space of global search, we embraced an evolutionary algorithm based on network growth for the search. With the intention to attest the effectiveness of our search methodology, we utilized the traditional EA method as a benchmark for the search by using One-Shot to train the model in the same global hypergraph. We executed a traditional EA with parameters identical to our methodology and searched directly for the comprehensive model. It was discerned that the top-1 accuracy of the model degenerated to 96.6\%, a decline of 0.9\% compared to G-EvoNAS. Due to this, it failed to parallel our search strategy. We ascertain that the performance discrepancy originates from the effectiveness of our network thriving method. In the G-EvoNAS approach, a single block is searched at one instance, and each architecture is evaluated in batches of a size equal to 512. Subsequently, the mean accuracy and average model size are quantified to symbolize the potential of each block. In contrast, the classic EA method necessitates simultaneous search of all blocks and the performance of the entire network must be evaluated. This leads to reduced search efficacy, especially in a large search space. Seeking the globally optimal solution demands superior performance of the search algorithm. As visible in Figure 4, the efficacy of our novel evolutionary search methodology is clearly demonstrated.


\begin{figure}[htbp]    
  \centering
\subfloat[]   
  {
      \label{fig:subfig-a}\includegraphics[width=0.23\textwidth]{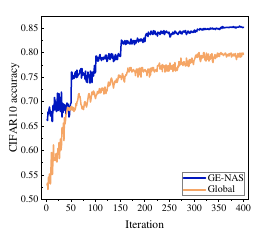}
  }
  \subfloat[]
  {
      \label{fig:subfig-b}\includegraphics[width=0.23\textwidth]{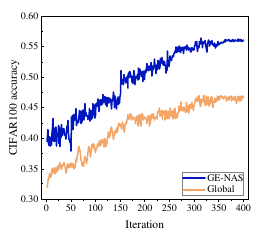}
  }
  \subfloat[]
  {
      \label{fig:subfig-c}\includegraphics[width=0.23\textwidth]{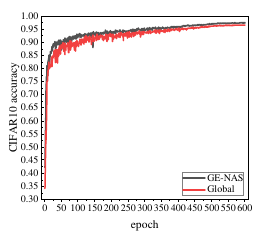}
  }
  \subfloat[]
  {
      \label{fig:subfig-d}\includegraphics[width=0.23\textwidth]{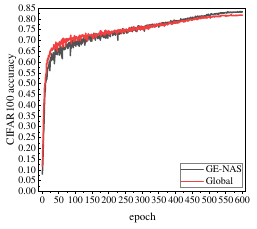}
  }
  \caption{Comparing the effectiveness of G-EvoNAS using network growing and SuperNet pruning on CIFAR10 and CIFAR100. In (a) and (b), the orange line shows the best fitness in each generation during the evolution iteration. The blue line marks the best validation accuracy of the models that were sampled by the  global search in each iteration of the parameter sharing method. In (c) and (d), we plot the validation accuracy of the two networks found by both methods during the training. Each network is trained for 600 epochs.}    
  \label{fig:subfig_2} 
\end{figure}

\noindent {\bf Effectiveness of Supernet Pruning} A distinct advantage of the One-Shot technique is that the shared weights of the hypergraph can be employed to conveniently predict the performance of various architectures. Nonetheless, earlier models \cite{guo2020single,liu2018progressive} exhibit inadequate results when ranking models. To ascertain the degree to which SuperNet pruning can mitigate this issue, we performed the following comparison. Firstly, we select a bunch of models from the individuals in the final population under a partitioned search space, train them from scratch, and evaluate their independent top-1 accuracy. Subsequently, we engage One-Shot to traditionally train the hypergraph beneath a block search space. Finally, we present the model rankings of G-EvoNAS and One-Shot using the accuracy obtained by model inference in hypergraphs trained by both methods. The divergence is visible in Figure 3 (b). The Kendall correlation coefficients for One-Shot and G-EvoNAS stand at 0.38 and 0.87 respectively. Hence, the models under G-EvoNAS SuperNet pruning can be ordered based on the precision of shared weight evaluation, which is visibly superior compared to that of One-Shot. Repeated experiments also unveiled analogous phenomena.

\section{Conclusion}
\label{sec:conclusion}
The Neural Architecture Search (NAS) holds the capacity to autonomously devise high-performance models. Diverging from preceding approaches that merely search local architectures, this paper proposes a block search space founded on the global scale. Simultaneously, with the objective to effectively explore superior models in the block search space, we propose an evolutionary neural architecture search based on network growth. Throughout this evolutionary journey, G-EvoNAS optimally harnesses the knowledge garnered in the latest generational evolution, specific to architecture and parameters. While utilizing pruned SuperNets to augment individual evaluation efficiency and maintain individual ranking consistency, experiments conducted on benchmark datasets reveal that the proposed G-EvoNAS can identify models within a shorter timeline, and with superior performance in model size/latency and accuracy parameters. It outstrips the performance of existing state-of-the-art models.

\clearpage  

%
%
\bibliographystyle{splncs04}
\bibliography{main}

\end{sloppypar}
\end{document}